%% file: sample-manuscript.tex
\documentclass[acmsmall]{acmart}
\renewcommand\footnotetextcopyrightpermission[1]{}
\AtBeginDocument{%
  \providecommand\BibTeX{{%
    \normalfont B\kern-0.5em{\scshape i\kern-0.25em b}\kern-0.8em\TeX}}}

\setcopyright{acmlicensed}
\copyrightyear{2018}
\acmYear{2018}
\acmDOI{XXXXXXX.XXXXXXX}

\acmConference[Conference acronym 'XX]{Make sure to enter the correct
  conference title from your rights confirmation emai}{June 03--05,
  2018}{Woodstock, NY}
\acmISBN{978-1-4503-XXXX-X/18/06}

\begin{document}

\def\mohit #1{\textcolor{red}{mohit: #1}}
\def\mdc #1{\textcolor{blue}{mdc: #1}}
\def\misinfo {\textit{Misinformative}}
\def\nonmisinfo {\textit{Non-Misinformative}}
\def\control {Control}
\def\treatment {Treatment}
\def\pericovid {\textit{Peri-COVID}}
\def\precovid {\textit{Pre-COVID}}
\def\climate {climate change}
\def\politics {politics}
\def\pval {pval}
\def\answerTODO #1{#1}
\definecolor{Maroon}{HTML}{AF3235}
\definecolor{Violet}{HTML}{58429B}
\definecolor{PineGreen}{HTML}{008B72}

\title[Understanding the Humans Behind Online Misinformation]{Understanding the Humans Behind Online Misinformation:\\ An Observational Study Through the Lens of the COVID-19 Pandemic}

\author{Mohit Chandra}
\affiliation{%
  \institution{School of Interactive Computing, Georgia Institute of Technology}
  \city{Atlanta, GA}
  \country{USA}}
\email{mchandra9@gatech.edu}

\author{Anush Mattapalli}
\affiliation{%
  \institution{School of Computer Science, Georgia Institute of Technology}
  \city{Atlanta, GA}
  \country{USA}}
\email{amattapalli3@gatech.edu}

\author{Munmun De Choudhury}
\affiliation{%
  \institution{School of Interactive Computing, Georgia Institute of Technology}
  \city{Atlanta, GA}
  \country{USA}}
\email{munmun.choudhury@cc.gatech.edu}

\renewcommand{\shortauthors}{Chandra, et al.}

\begin{abstract}

The proliferation of online misinformation has emerged as one of the biggest threats to society. 
Considerable efforts have focused on building misinformation detection models, still the perils of misinformation remain abound. Mitigating online misinformation and its ramifications requires a holistic approach that encompasses not only an understanding of its intricate landscape in relation to the complex issue- and topic-rich information ecosystem online, but also the psychological drivers of individuals behind it. 
Adopting a time series analytic technique and robust causal inference-based design, we conduct a large-scale observational study analyzing over 32 million COVID-19 tweets and 16 million historical timeline tweets. We focus on understanding the behavior and psychology of users disseminating misinformation during COVID-19 and its relationship with the historical inclinations towards sharing misinformation on Non-COVID domains before the pandemic. Our analysis underscores the intricacies inherent to cross-domain misinformation, and highlights that users' historical inclination toward sharing misinformation is positively associated with their 
present behavior pertaining to misinformation sharing on emergent topics and beyond. This work may serve as a valuable foundation for designing user-centric inoculation strategies and ecologically-grounded agile interventions for effectively tackling online misinformation.
\end{abstract}



\received{20 February 2007}
\received[revised]{12 March 2009}
\received[accepted]{5 June 2009}

\maketitle

\section{Introduction}
\label{sec:introduction}
\input{sections/introduction}

\section{Related Work}
\label{sec:related_work}
\input{sections/related_works}

\section{Data Collection and Classification}
\label{sec:data_collection_clasification}
\input{sections/data_collection_pipeline}

\section{RQ1: User Behavior During COVID-19}
\label{sec:rq1}
\input{sections/rq_1}

\section{RQ2: Impact of Past Behavior on Sharing Misinformation During COVID-19}
\label{sec:rq2}
\input{sections/rq2_temporal_analysis}

\section{RQ3: Impact of COVID-19 on Users' Psychological Behaviors}
\label{sec:rq3}
\input{sections/rq3_liwc_analysis}

\section{Discussion}
\label{sec:discussion}
\input{sections/discussion}

\bibliographystyle{ACM-Reference-Format}
\bibliography{sample-base}

\end{document}

%% file: sections/introduction.tex
\begin{quote}
    ``We don’t fall for false news just because we’re dumb. Often it’s a matter of letting the wrong impulses take over.'' -- TIME \cite{timemagazinemisinfo}
\end{quote}

Historian Yuval Noah Harari, in his celebrated book, {\it 21 Lessons for the 21st Century} argued that ``fake news is much older than Facebook''~\cite{harari201821}. 
In fact, myths have served many purposes in defining the human civilization as we know of today -- from the dawn of the Stone Age onward, self-perpetuating myths have played a vital role in binding together human communities. 
However, a delicate balance between truth and fiction defines when myths pave the way for human cooperation, and when they predispose us to social unrest, violence, and polarization, or contribute to eroding trust in institutions. 
With the increasing adoption of social media as the predominant source of news consumption for a substantial segment of the populace~\cite{pewresearchsurvey2021}, this delicate balance is tested everyday. 

Homo sapiens may have always been a ``post-truth species'' ~\cite{harari2018yuval}, but the 2020 COVID-19 pandemic could be viewed as an inflexion in the recognition of the threats of online misinformation. During this global crisis, the world not only struggled with the physical and social implications of the virus but also found itself in the grasp of an infodemic~\cite{PIAN2021102713}. Misinformation during this time had varied ramifications such as health hazards/deaths~\cite{islam2020covid} due to untested treatment methods, exacerbated racism and hate-speech~\cite{10.1145/3465336.3475111}, worsening mental health
~\cite{verma2022examining} to name a few. 
Consequently, the past few years have witnessed burgeoning investments  in detecting misinformation on social media, primarily focused on building automated or semi-automated machine learning models to support platform moderation efforts~\cite{li2022youtube, kouzycovid}. 

However, the COVID-19 infodemic~\cite{PIAN2021102713} was not the society's first encounter with widespread online misinformation, and it certainly will not be the last. Anti-vax, scientific, political, and climate change related misinformation have been around for a while, defining our post-truth societies, only to have gained momentum in the social media era. Tackling the perils of online misinformation requires a holistic approach that looks at the ecology of such multifaceted online misinformation, rather than a ``whack-a-mole" approach, where we address one issue at a time as it arises~\cite{ruchansky2017csi}. Misinformation is adaptive and evolving~\cite{jaeger2021arsenals}, and by the time one aspect is addressed, it may have already shifted elsewhere. Moreover, misinformation is complex -- it permeates various aspects of our lives, from politics and health to science and culture. 
Unraveling the intricate interconnectedness between the sharing practices of various misinformation is key to mitigation efforts that are synergistic across situations and topics.  

Further, to curb the threats of online misinformation, we have to unpack the underlying psychologies and identify key driving mechanisms. In turn, this implies delving into understanding the \textit{humans} behind misinformation. 
As rightly captured in the opening quote from the TIME magazine, an assumption that misinformation sharing is solely attributed to ignorance is far too simplistic. It may stem from confirmation bias, a lack of media literacy, personal gain, ideological reasons, or simply because of one's vulnerability to sensationalized information~\cite{ecker2022psychological}. Although studies have shown that automated detection systems demonstrate greater efficacy when applied to user-level data as opposed to content-level data~\cite{qian-etal-2018-leveraging}, to our knowledge, a comprehensive understanding of the users (authors of misinformative posts) is lacking. Importantly, prior research has demonstrated that a limited proportion of users are responsible for generating the majority of misinformation within the digital sphere~\cite{ccdhreport}. Therefore, mitigation strategies directed at the user can be considered a pragmatic approach, rather than content or user moderation (deletion, banning, censoring) that could be perceived to be vilifying individuals or impediments to free speech.

To this end, we adopt a tri-prong approach looking at the interrelationships between historical sharing of misinformation relating to diverse domains (\politics~and \climate~in particular), the occurrence of a global crisis (the COVID-19 pandemic), and subsequent sharing of misinformation on new domains (COVID-19) by a panel of nearly 6 thousand Twitter users. Adopting a time series analytic technique and robust causal inference based design, we analyzed over 32 million tweets related to COVID-19 posted during the first 5 months of 2020 and over 16.5 million historical timeline tweets posted over a period of 17 months. 

This paper addresses the following research questions: 

\vspace{0.04in} \noindent\textbf{RQ1:} \textit{How do posting behaviors of users who share COVID-19 misinformation and those who do not differ?}

\vspace{0.04in} \noindent\textbf{RQ2:} \textit{How does historical sharing of online misinformation on various domains (\politics~and \climate) impact people's inclination towards disseminating online misinformation on other domains (COVID-19) in the future?} 

\vspace{0.04in} \noindent\textbf{RQ3:} \textit{How did the COVID-19 pandemic impact and alter the psychological behavior of the users who historically shared or did not share misinformation on other domains (\politics~and \climate)?}

\vspace{0.04in} 
First, per \textbf{RQ1}, contrasting the posting behaviors of COVID-19 related posts among the \misinfo~users and their \nonmisinfo~counterparts, we find that 
\misinfo~users' behavior is influenced by US politics and racial prejudice, while \nonmisinfo~users focus on COVID-19 news, treatments, and fatalities. Furthermore, \misinfo~users initiate and sustain COVID-19 discussions longer than their counterparts. For~\textbf{RQ2}, we employ a causal analysis framework on 5,938 users, divided into \control~and \treatment~groups based on their historical misinformation-sharing activity. We find that approximately 70\% of \treatment~users are classified as COVID-19 \misinfo~users, while about 84\% of \control~users are COVID-19 \nonmisinfo~users. Furthermore, \treatment~users show higher engagement in COVID-19 and Non-COVID (\politics, \climate) misinformation dissemination during the pandemic. Finally, for \textbf{RQ3}, 
we employ Difference-in-Difference analysis~\cite{angrist2009mostly} studying the psycholinguistic characteristics of tweets unrelated to COVID-19. We note a ``spillover'' effect of the pandemic, with \treatment~users showing signs of potential misinformation spreading in their language use, while the \control~group adopting a more measured approach in their general social media discussions.

Our findings underscore the existence of an intricate association between users' historical misinformative behaviors online and their present actions on social media, surrounding misinformation sharing as well as beyond. Hence, the strategic application of user-centric interventions targeted at these identified users as well as inoculation efforts emerge as both pragmatic as well as preemptive 
 approaches for curtailing the propagation of misinformation in forthcoming contexts. 
Further, our study 
contributes to understanding the complex interplay between misinformation concerning various domains during the course of a global crisis. 
It thus emphasizes the imperative for stakeholders to engage in the study of cross-domain misinformation to understand its impact on people and the online information ecosystem holistically. 


\paragraph{Privacy and Ethics.} 
Although we use public social media data, we worked with deidentified data and refrained from sharing raw or personally identifiable data in any form. 
Since we analyzed retrospective data and did not interact with the authors of tweets, our Institutional Review Board did not consider it human subjects research and therefore did not require IRB approval. Still, we adopted best practices given in the literature on social media data analysis~\cite{weller2016manifesto,weller2015uncovering}. 
Notably, we only report aggregated results for users, ensuring that we are not exposing any personally identifiable information of any user. Following Twitter's data-sharing guidelines and appropriate data-use agreements, we intend to make available the curated dataset exclusively comprising Tweet IDs, to interested researchers of online misinformation. 
Consequently, any curated dataset shared externally will not contain any direct association with personally identifiable information of users. 

%% file: sections/related_works.tex

\subsection{Detection and analysis of online misinformation}

The detection and trend analysis of misinformation across diverse domains on social media has garnered substantial scholarly attention in recent times. 
Focusing on COVID-19 misinformation, studies have focused on various topics, such as creating datasets, and social media content analysis. \citet{cui2020coaid} introduced the CoAID dataset that included over 4,200 news articles, 296,000 related user engagements, and 926 social platform posts about COVID-19. In another study, \citet{micallef2020role} curated a dataset of false claims and statements for counter-misinformation on fake cures and 5G conspiracy theories related to COVID-19. In a later study, \citet{10.1145/3465336.3475111} presented the CoronaBias dataset containing over 440,000 tweets focusing on Islamophobia during COVID-19 and performed a longitudinal study. On a related tangent, content analysis on different social media platforms has attracted a large number of studies. In one of the early studies,~\citet{kouzycovid} analyzed the correlation between different accounts (individual, public figures) and hashtags with the rate of misinformation spread. \citet{li2022youtube} on the other hand, analyzed Youtube videos and found that over one-quarter of the most viewed YouTube videos on COVID-19 contained misleading information during March 2020. 

Recently, political misinformation and fake news has drawn attention from researchers, especially dissemination of misinformation during elections has been studied extensively.~\citet{article2016politicsfakenews} found that 25\% of tweets sharing news links during the 2016 US presidential elections promoted fake news. A few studies have also focused on creating datasets and classifiers for detecting political misinformation.~\citet{wang2017liar} presented the LIAR dataset comprising 12.8K manually labeled short statements collected from PolitiFact.com. Another noted study proposed FakeNewsNet, a repository containing tweets related to fake news debunked via PolitiFact and Gossip Cop~\cite{shu2020fakenewsnet}. In contrast, misinformation detection pertaining \climate~is relatively under-studied. Only a few studies include datasets focusing on climate change misinformation,~\citet{bhatia2020right} created a dataset comprising 1,168 articles focusing on climate change-related misinformation. In a later study, \citet{diggelmann2021climatefever} proposed CLIMATE-FEVER dataset comprising 1,535 real-world claims on climate change.

Prior research has examined COVID-19, \politics, and \climate~misinformation separately, but this compartmentalized approach does not mirror real-world complexity, where users often encounter misinformation covering multiple topics. In contrast, our study delves into cross-domain misinformation, revealing the intricate interplay among various domains during the pandemic.

\subsection{User-Centric analysis of misinformation}

One of the major focuses of interest in the work related to COVID-19 misinformation was on studying the implications of COVID-19 misinformation on various aspects related to society. In one of the popular studies, \citet{ROMER2020113356} observed that the belief in conspiracy theories prospectively predicted resistance to preventive action and vaccination among individuals.~\citet{islam2020covid} showed that COVID-19 misinformation fueled by rumors, stigma, and conspiracy theories had serious implications on the individual and community, especially on public health. In a later study, \citet{verma2022examining} demonstrated that users who shared COVID-19 misinformation experienced approximately two times an additional increase in anxiety when compared to similar users who did not share misinformation. Cases of hate speech against specific communities surged during the pandemic due to the origin theories related to the virus. Focusing on the anti-Asian hate, \citet{10.1145/3487351.3488324} performed a longitudinal analysis studying the patterns among users spreading anti-Asian hate and those countering it.

Compared to COVID-19, user-centric analysis pertaining to \politics~and \climate~misinformation is under-studied. \citet{osmundsen2021partisan} analyzed 2,300 American Twitter users and observed that strong support for partisan polarization was an influential driver for sharing political misinformation. In one of the popular works in \climate~misinformation, \citet{van2017inoculating} explored how people evaluate and process consensus cues in a polarized information environment pertaining to climate change-related topics. Later, \citet{benegal2018correcting} showed that publicizing the views of influential Republicans acknowledging scientific consensus on climate change can reduce the partisan gap.

Unlike prior studies, we examine the causal relationship between users' past and future engagement with COVID-19 and non-COVID (\politics~and \climate) misinformation. Furthermore, we study the impact of the pandemic on altering the psychological behavior of the users.

%% file: sections/data_collection_pipeline.tex
\begin{figure*}[!t]
    \centering
        \includegraphics[width=\columnwidth]{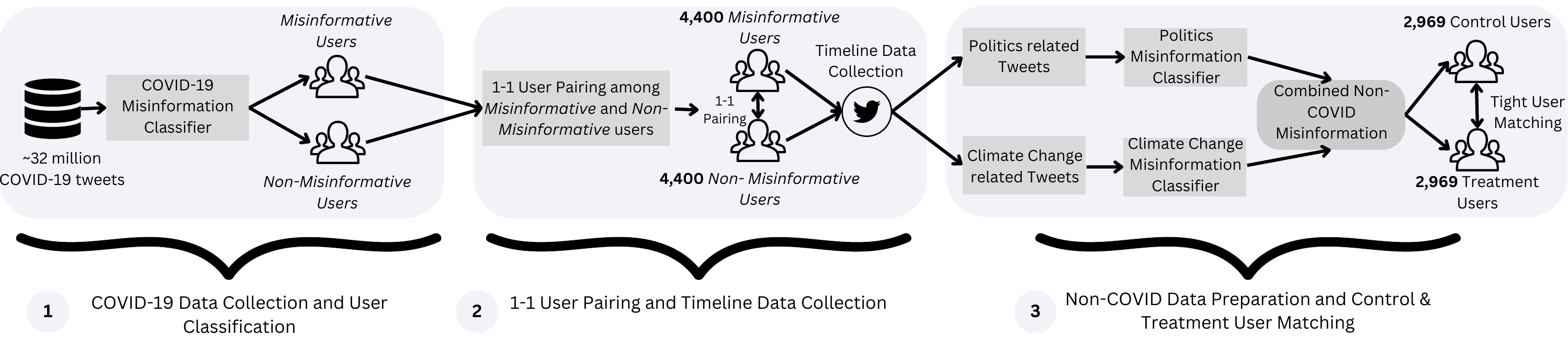}
    \caption{An overview figure summarizing the data collection methodology for our study. The pipeline has three main components pertaining to data collection and user classification. The number and text below each part explain the method.} 
    \label{fig:data_pipeline}
\end{figure*}

We begin by describing our data collection approach, along with our methodology for processing this data towards answering the RQs. Figure~\ref{fig:data_pipeline} gives an overview of our strategy (Phases 1-3, left to right), each of which is elaborated below.

\subsection{Phase 1: COVID-19 Data Collection; Misinformation Detection; User Classification}
We referred to~\citet{info:doi/10.2196/19273}'s COVID-19-TweetIDs dataset that contains IDs for posts containing keywords and hashtags related to the pandemic spanning between $21^{st}$ Jan 2020 and $17^{th}$ Feb 2023. 
Given the longitudinal nature of our study and an emphasis on the initial part of the pandemic, we hydrated the tweet IDs, 
using the Twitter API, 
to obtain $\sim32.9$ million tweets along with metadata between $21^{st}$ Jan 2020 and $31^{st}$ May 2020. 

Next, we used BERTweet COVID-19 Uncased~\cite{nguyen-etal-2020-bertweet}
~as the base classifier to identify COVID-19 misinformation in this collected data. 
BERTweet model has been pre-trained on five million COVID-19-related tweets and 845 million tweets related to other topics. We fine-tuned the model on the combination of two publicly available datasets datasets -- 1) CoAID dataset~\cite{cui2020coaid}, 2) COVID cure and 5G dataset~\cite{micallef2020role}. We fine-tuned the model with a learning rate of 2$e^{-5}$ for a maximum of 8 epochs using 18,316 examples, out of which 15,260 were labeled as non-misinformative and 3,056 as misinformative.
We also used Batch Normalization and Dropout to avoid over-fitting and improve stability. Table~\ref{tab:classifier_performance} presents the performance metrics for the fine-tuned classifier across 10-fold cross-validation. We used the fine-tuned COVID-19 misinformation classifier to label $\sim32.9$ million COVID-19 tweets. Additionally, we kept the threshold of $\geq0.9$ for a high-precision misinformation classification of COVID-19 tweets. Table~\ref{tab:data_stats} presents the statistics related to this misinformation classification.

\textbf{RQ1} analyzes the posting behavior differences among the \textit{Misinformative} and \textit{Non-Misinformative} users. Hence, after classifying the tweets into one of the two categories (misinformation or not) pertaining to COVID-19, we used an empirically-driven strategy to assign each user a label as \textit{Misinformative} or \textit{Non-Misinformative}. We use a combination of past work~\cite{verma2022examining} and empirical statistics to define the threshold. Specifically, we assign the \textit{Misinformative} label to a user if -- 1) the user has five or more COVID-19 misinformation tweets~\cite{verma2022examining}, and 2) the ratio of \#misinformation tweets to total tweets by the user is equal or greater than 0.15. The latter condition was added empirically from data statistics to counter the chances of misclassification for users tweeting with high frequency. Using this methodology, we classified 162,726 users as \misinfo~and 7,221,301 users as \nonmisinfo. 

\begin{table}[ht]
    \centering
    \footnotesize
    \begin{tabular}{|l|p{0.12\columnwidth}|p{0.12\columnwidth}|p{0.12\columnwidth}|p{0.12\columnwidth}|}
    \hline
         \textbf{Classifier} & \textbf{Acc.} & \textbf{Recall} & \textbf{Precision} & \textbf{F-1} \\
         \hline
         COVID-19  & $0.937\pm0.017$ & $0.921\pm0.011$ & $0.874\pm0.030$ & $0.894\pm0.023$ \\
         \hline
         Politics  & $0.842\pm0.009$ & $0.847\pm0.012$ & $0.838\pm0.010$ & $0.839\pm0.009$ \\
         \hline
         Climate Change & $0.850\pm0.031$ & $0.846\pm0.036$ & $ 0.831\pm0.033$ & $0.835\pm0.034$ \\
         \hline
    \end{tabular}
    \caption{Performance metrics of the three misinformation classifiers across 10-fold cross-validation.}
    \label{tab:classifier_performance}
    \vspace{-1.5em}
\end{table}

\begin{table}[ht]
    \centering
    \footnotesize
    \begin{tabular}{|p{0.3\columnwidth}|p{0.25\columnwidth}|p{0.25\columnwidth}|}
    \hline
         Data Type & \textbf{Total Tweets} & \textbf{Misinformation Tweets} \\
         \hline
         COVID-19 & 32,922,320 & 5,363,208 \\
         \hline
         Politics~\& Climate Change & 6,221,419 & 1,913,254 \\
         \hline
    \end{tabular}
    \caption{Data Statistics.} 
    \label{tab:data_stats}
    \vspace{-1.5em}
\end{table}


\subsection{Phase 2: User Pairing; Timeline Data Collection} 

For \textbf{RQ2} and \textbf{RQ3} we also study the relationship between \precovid~and \pericovid~behavior among users. Therefore, corresponding to Phase 2 in Figure~\ref{fig:data_pipeline}, we collected timeline data for users from $1^{st}$ January 2019 till $31^{st}$ May 2020. Due to the limitation of the monthly tweet cap (10 million tweets per month) of the then Academic Twitter API, we could collect the timeline tweets for a subset of the users. We followed a process for creating 1-1 pairs among the users from \misinfo~and \nonmisinfo~user groups to eliminate the possibility of any confounding variables and to ensure similarity among the users. 
We randomly sampled 15,000 \misinfo~users and 200,000 \nonmisinfo~users. As the next step, for each user, we computed the co-variates that involved -- 1) user bio statistics (follower count, following count, account creation date), 2) consolidated user activity (avg. status count, favourite count), and 3) month-wise user activity (like and retweet count for each month from Jan-May 2020). Next, we used a Logistic Regression classifier to predict each user's propensity score based on the co-variates mentioned. Finally, we paired each \misinfo~user with the most similar \nonmisinfo~user using the K-nearest neighbor algorithm, keeping a caliper distance $\leq$ 0.2
~\cite{Kiciman_Counts_Gasser_2018}. Thus we obtained a subset of 8,800 users (4,400 from each set). We were able to collect timeline data for 4,379 \misinfo~users and 4,397 \nonmisinfo~users. In total, we collected $\sim$29.2 million timeline tweets at the end of Phase 2. 

\subsection{Phase 3: Non-COVID-19 Data Collection and Misinformation Detection} 

In addition to analyzing the COVID-19-related tweets, our study also focuses on Non-COVID-19-related misinformation. While various aspects can be covered in this area, we specifically focus on \climate~and \politics~related misinformation, 
due to their relatively more comprehensive treatment within the literature 
~\cite{shu2020fakenewsnet, diggelmann2021climatefever}. We curated several high-precision keywords related to \climate~and \politics~from past works and reliable sources~\cite{EPA_2016, The_Britannica_Dictionary, politics_wikipedia}.
We used this list to filter data related to \climate~and \politics~from the collected timeline tweets. Finally, we obtained $135,952$ \climate~related and $6,085,467$ \politics~related tweets in this phase.  


Then, for the \politics~misinformation classifier, we used PoliBERTweet trained on 83 million tweets related to politics.
~We fine-tuned the classifier on two publicly available politics misinformation datasets -- 1) Liar~\cite{wang2017liar}, and 2) FakeNewsNet~\cite{shu2020fakenewsnet}. 
We used 6,962 examples, of which 4186 were non-misinformative and 2776 were political misinformation. For the \climate~misinformation classifier, we used ClimateBERT
~pre-trained on 2+ million paragraphs of climate-related topics. We fine-tuned the \climate~classifier on two public datasets -- 1) Climate-Fever~\cite{diggelmann2021climatefever}, and 2) CCCM~\cite{bhatia2020right}. We combined the data and augmented the data using back translation, 
resulting in 3,903 examples of which 1,301 were non-misinformative and 631~\climate~misinformative examples. 
Each classifier was fine-tuned with a learning rate of 2$e^{-5}$ for a maximum of 8 epochs. Additionally, we used Batch Normalization and Dropout in each classifier to avoid over-fitting and improve stability. Table~\ref{tab:classifier_performance} shows the performance of these classifiers 
across 10-fold cross-validation. 

In the next step, we used the fine-tuned \politics~and \climate~classifiers to label the Non-COVID-19 data. We labeled $\sim6.09$ million \politics~related tweets using the \politics~ misinformation classifier and $\sim136$ thousand \climate~related tweets using the \climate~misinformation classifier 
For each classifier, we kept the threshold of $\geq0.9$ for a high-precision misinformation classification of tweets for each type. 
~Finally, the classified \politics~and \climate~tweets were subsequently used in a causal inference-based statistical matching setup, described in forthcoming Section 5 on \textbf{RQ2}.

%% file: sections/rq_1.tex
External stimuli can lead to internal reactions in individuals, shaping their behavior. S-O-R (Stimulus-Organism-Response), a popular framework to study this phenomenon, underscores the importance of individuals’ psychological processes in response to external stimuli~\cite{hsiao2021captures}. Inspired from~\citet{laato2020unusual}, we adopted the S-O-R framework to investigate how real-world events that happened during the \pericovid~era affected the posting behavior and user activity among the \misinfo~and \nonmisinfo~users. We analyze two aspects here -- 1) relationship between posting trends and external event change points, 2) user inception and continuation.

\begin{table*}[t]
\centering
\footnotesize
\setlength{\tabcolsep}{4pt}
\begin{tabular}{|m{0.1\textwidth}|m{0.35\textwidth}|m{0.48\textwidth}|}
\hline
\textbf{Date} & \textbf{External Events} & \textbf{Topics}\\
\hline \hline 
\rowcolor{red!5} \multicolumn{3}{|c|}{\textbf{Misinformative Users}} \\
\hline \hline 
2020-01-31 &  Wuhan lockdown announced, UN calls emergency meetings for COVID-19. & \begin{itemize}
    \item \textcolor{Maroon}{COVID-19 declared as a public health emergency.}
    \item \textcolor{Violet}{Racism against the Chinese Communist Party for COVID-19.}
    \item \textcolor{PineGreen}{Discussions on political issues involving Putin, corruption, and hacking of US elections.}
\end{itemize} \\
\hline
2020-02-24 & Italy and Iran became hotspots for COVID-19 cases & \begin{itemize}
    \item \textcolor{Maroon}{Blaming Trump for a weak response to COVID-19.}
    \item \textcolor{Violet}{Increased number of deaths in Iran due to Coronavirus.}
    \item \textcolor{PineGreen}{Large number of cases of COVID-19 in Lombardy region of Italy.}
\end{itemize} \\
\hline
2020-03-13 & COVID-19 declared a pandemic, and 8 US states declared a state of emergency. & \begin{itemize}
    \item \textcolor{Maroon}{Users discussing the racist propaganda on the origin of COVID-19 in US or China.}
    \item \textcolor{Violet}{Suggestions for testing any flu-like symptoms for COVID-19.}
    \item \textcolor{PineGreen}{Sick patients being denied COVID-19 tests across the US.}
\end{itemize} \\
\hline
2020-03-23 & Donald Trump invokes the Defense Production Act and WHO announces that the pandemic is accelerating. & \begin{itemize}
    \item \textcolor{Maroon}{Racist tweets calling COVID-19 as Chinese Virus.}
    \item \textcolor{Violet}{Discussion related to the importance of social distancing.}
    \item \textcolor{PineGreen}{People questioning giving up their liberty for COVID-19, and ignorance related to COVID-19.}
\end{itemize} \\
\hline
2020-03-24 & Donald Trump invokes the Defense Production Act and WHO announces that the pandemic is accelerating. & \begin{itemize}
    \item \textcolor{Maroon}{Discussion on social distancing measures related to COVID-19.}
    \item \textcolor{Violet}{Calling out Trump's failure in handling COVID-19.}
    \item \textcolor{PineGreen}{Man dies after ingesting Chloroquine in an attempt to prevent COVID-19.}
\end{itemize} \\
\hline
\end{tabular}
\caption{Change point Topics for Misinformative Users. The three most popular topics among the posts from Misinformative users per event are shown.}
\label{tab:rq_1_misinfo_changepoints}
\vspace{-1.5em}
\end{table*}

\begin{table*}[t]
\centering
\footnotesize
\setlength{\tabcolsep}{4pt}
\begin{tabular}{|m{0.1\textwidth}|m{0.35\textwidth}|m{0.48\textwidth}|}
\hline
\textbf{Date} & \textbf{External Events} & \textbf{Topics}\\
\hline \hline 
\rowcolor{green!5} \multicolumn{3}{|c|}{\textbf{Non-Misinformative Users}} \\
\hline \hline 
2020-01-31 & Wuhan lockdown announced, UN calls emergency meetings for COVID-19. & \begin{itemize}
    \item \textcolor{Maroon}{Precautionary measures for COVID-19.}
    \item \textcolor{Violet}{COVID-19 declared a global health emergency.}
    \item \textcolor{PineGreen}{Brexit came into effect in the UK.}
\end{itemize} \\
\hline
2020-03-02 & Rising COVID-19 cases in several countries, including Italy. & \begin{itemize}
    \item \textcolor{Maroon}{Using sanitizers, face masks and gloves to protect against COVID-19.}
    \item \textcolor{Violet}{Stocks market going down due to fears revolving around COVID-19.}
    \item \textcolor{PineGreen}{Rise in COVID-19 cases and death tolls in Italy.}
\end{itemize} \\
\hline
2020-03-07 & FDA approves the EUA for CDC's SARS-CoV-2 diagnostic test kit and rising cases in Italy. & \begin{itemize}
    \item \textcolor{Maroon}{Rising cases and high numbers of deaths in Italy's Lombardy region.}
    \item \textcolor{Violet}{Tips for avoiding the spread of COVID-19.}
    \item \textcolor{PineGreen}{Toilet paper crisis due to COVID-19 in the US.}
\end{itemize} \\
\hline
2020-03-13 & COVID-19 declared a pandemic, and 8 US states declared a state of emergency. & \begin{itemize}
    \item \textcolor{Maroon}{Rescheduling and cancellation of local and global events due to COVID-19.}
    \item \textcolor{Violet}{Closing of schools in various states across the US.}
    \item \textcolor{PineGreen}{Following the measures suggested by medical experts.}
\end{itemize} \\
\hline
2020-05-20 & President Xi Jinping and President Donald Trump announcing plans about COVID-19 vaccine. & \begin{itemize}
    \item \textcolor{Maroon}{Latest news and updates related to COVID-19 from various countries worldwide.}
    \item \textcolor{Violet}{Importance of following the precautionary measures against COVID-19.}
    \item \textcolor{PineGreen}{Racist comments related to the origin of COVID-19.}
\end{itemize} \\
\hline
\end{tabular}
\caption{Change point Topics for Non-Misinformative Users. The three most popular topics among the posts from Non-Misinformative users per event are shown.}
\label{tab:rq_1_non_misinfo_changepoints}
\vspace{-1.5em}
\end{table*}

\subsection{Temporal Change Point Analysis}

One of the prevalent approaches for discerning significant trends and occurrences involves the examination of peak activities within the realm of social media, and subsequently correlating these peaks with events in the offline world~\cite{10.1145/3487351.3488324,micallef2020role}. Although this methodology provides valuable insights into the events that cause heightened interest on social media platforms, it tends to overlook occurrences that, while not creating spikes, exert discernible influences on user engagement. To overcome this issue, we used change point analysis to study the induction relationship between real-world events and the posting behavior of the two user sets. Change point analysis aims to identify instances of a shift in the underlying probability distribution of a time series. Moreover, the change point analysis procedure
is generalizable and can be performed on all types of time-ordered data. For this analysis, we used the Pruned Exact Linear Time (PELT) algorithm~\cite{killick2012optimal} to compute the changepoints. By maximizing the likelihood of the distribution given the data, the PELT algorithm detects the points in time at which the mean and variance of the distribution change.

We computed total tweets per day for the \misinfo~and \nonmisinfo~users and ran the PELT algorithm with a penalty over the tweet distributions of these two user sets. We ran the algorithm with penalty values 1-10 in decreasing order and added the change points on the first instance they occurred to provide a ranking among the change points (the greater the penalty, the higher the significance of the event). After obtaining change points for the two sets of users, we retrieved all unique tweets for the five most significant change point dates and used BERTopic~\cite{grootendorst2022bertopic}
to retrieve the most popular three topics. Through this, we analyzed the most popular events of interests among the~\misinfo~and~\nonmisinfo~users.

Table~\ref{tab:rq_1_misinfo_changepoints} and Table~\ref{tab:rq_1_non_misinfo_changepoints} 
presents the three most popular topics for the five most significant change point dates for \misinfo~users and \nonmisinfo~users respectively during the \pericovid~era. Interpretable topic labels were generated through manual inspection of representative terms in each topic, while the major event for each change point date was derived through consultation of prominent sources (e.g., WHO timeline, CDC)~\cite{whoevents2020,cdcevents2020,thinkglobalevents2020}. We observe two major contrasting patterns among the topics discussed among the user sets throughout the various change points. First, we observed topics highlighting the racist remarks targeting China majorly focused on creating/propagating conspiracy theories related to COVID-19 (such as calling COVID-19 as \textit{Chinese Virus}) being discussed 
among the \misinfo~users. These remarks majorly focused on creating/propagating conspiracy theories related \textit{Chinese Communist Party} involved in creating COVID-19. Such racist tweets were not found in the top topics discussed among the \nonmisinfo~users (except on the change point dated 2020-05-20). Second, in the case of \nonmisinfo~users majority of topics across all change points revolved around preventive measures for COVID-19 (such as using masks, sanitizers, social distancing) and discussion related to other aspects (such as the closure of institutions, lockdown etc.). In contrast, a few topics among the change points for the \misinfo~users were politically aligned, blaming the then US President Donald Trump for mishandling the COVID-19 response, cutting down CDC funding, etc. While these political topics may not be directly misinformative, these highlight the inclination of \misinfo~users towards political aspects of the pandemic, which is not highly visible among the other set. Additionally, as observed in Table~\ref{tab:rq_1_misinfo_changepoints} and Table~\ref{tab:rq_1_non_misinfo_changepoints}, for same change point dates (2020-01-31 and 2020-03-13) conspiracy theories related to the origin of the COVID-19 virus were discussed among the~\misinfo~users, whereas discussion among the~\nonmisinfo~users were centered around increasing cases and precautionary measures related to COVID-19.

\begin{figure*}[h!]
    \centering
    \includegraphics[width=0.47\columnwidth]{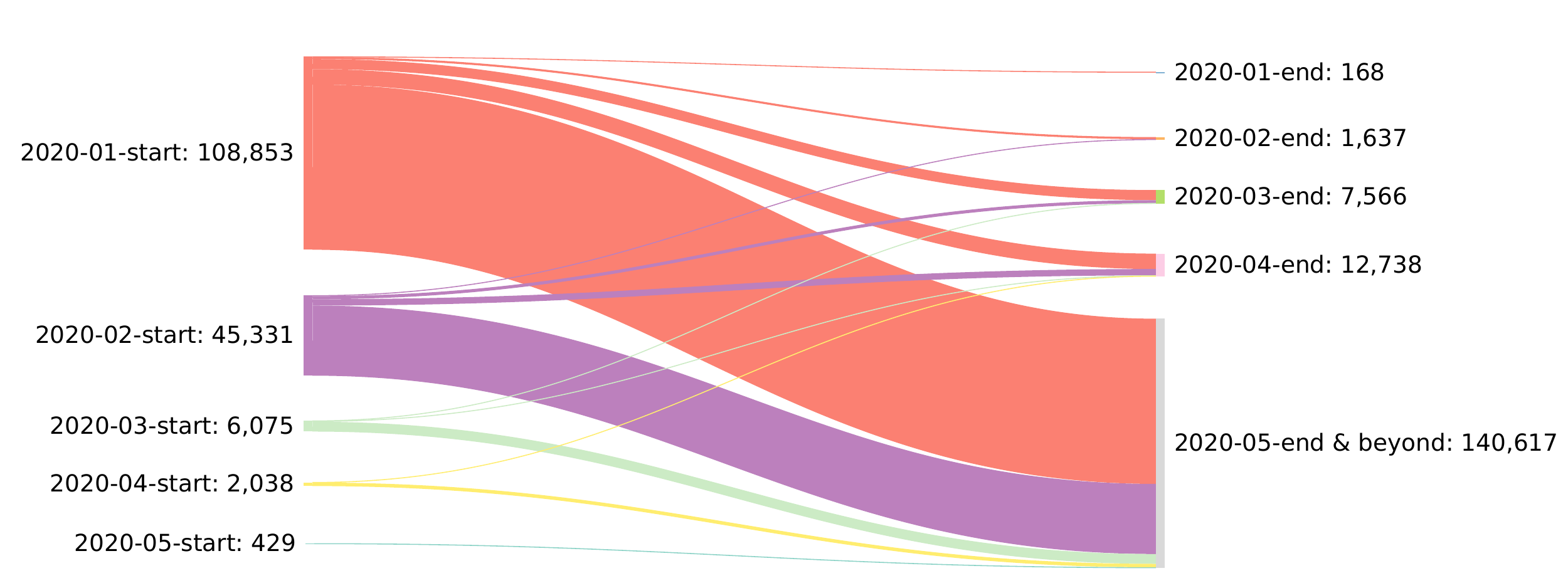}
    \includegraphics[width=0.47\columnwidth]{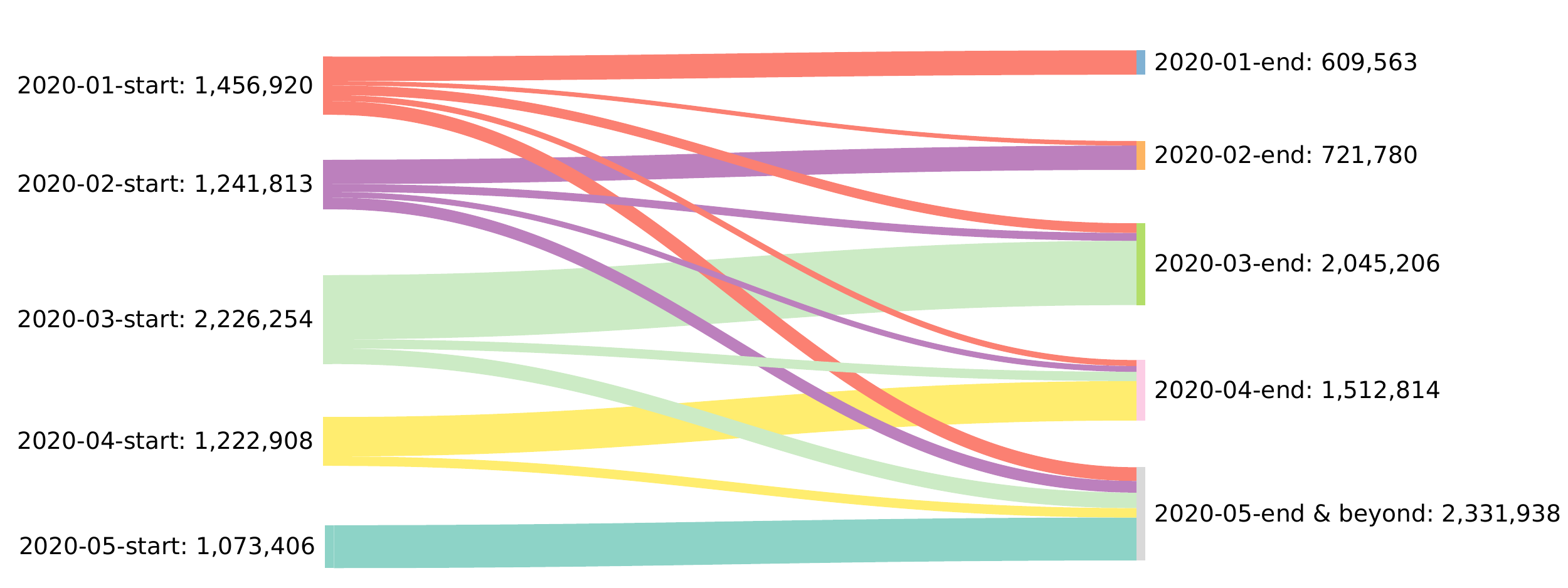}
    \caption{Sankey plots for the \misinfo~ (left) and \nonmisinfo~ users (right). Each plot presents the statistics related to users active (posted at least one tweet in the respective month) between specific ``starting'' and ``ending'' months.} 
    \label{fig:inception_sankey_plots_misinfo_non_misinfo}
    \vspace{-.15in}
\end{figure*}

\subsection{Inception and Continuation Analysis}

To further understand the influence of the COVID-19 pandemic on the posting behavior of \misinfo~and \nonmisinfo~users (per RQ1), we focused on their period of continued activity during the early months of COVID-19. Past works have shown that early detection of misinformation during significant events can help limiting it~\cite{ecker2022psychological}. Hence, we wanted to understand if the engagement among ~\misinfo~users differs from their counterpart during the early phase of COVID-19. While the temporal analysis revealed the constrasting nature of topics discussed among the~\misinfo~and~\nonmisinfo~users, the primary objective of this analysis was to ascertain whether the pandemic exerted varying effects on user engagement in discussions related to COVID-19 among the two groups.

For each user, we computed the month of their earliest tweet (referred as ``starting'' month) and the latest tweet related to COVID-19 (referred as ``ending'' month). For each pair of starting and ending month, we computed the number of users from both sets. Figure~\ref{fig:inception_sankey_plots_misinfo_non_misinfo} present the Sankey plots for the \misinfo~and \nonmisinfo~users respectively. We observe many contrasting differences between the two groups. While $66.89\%$ \misinfo~users tweeted their first tweet related to COVID-19 in January 2020, this ratio reduced to $27.86\%$ in February. On the other hand, we observe a more evenly distributed behavior among the \nonmisinfo~users with $30.83\%$ users starting to tweet about COVID-19 during March 2020, followed by $20.18\%$ during January. We further find a distinguishing pattern among the ``ending'' months between the two groups. $86.41\%$ of \misinfo~users continued posting about COVID-19 until May 2020 or beyond the dataset collection time period. However,  we observe a more evenly distributed percentage among the \nonmisinfo~users each month with $\sim32.29\%$ continued sharing COVID-19-related tweets until May 2020 or beyond.

The results from this analysis highlight the early adoption and 
behavior among the \misinfo~users where they jump on the domain (here COVID-19) early and keep discussing the topic for a longer duration as compared to their \nonmisinfo~counterparts.


%% file: sections/rq2_temporal_analysis.tex
Past research has shown that users who frequently disseminate misinformation exhibit resistance to change~\cite{ecker2022psychological}. Hence,  we aimed to ascertain whether this resistance to attitude change was observable in cases involving cross-domain misinformation. Specifically, in \textbf{RQ2} we investigate whether users' pre-pandemic online behavior significantly influenced their conduct during the \pericovid~period. We analyze the disparities in behavior between two distinct groups: the \control~group and the \treatment~group during the \pericovid~era pertaining to COVID-19 and Non-COVID (\politics~and \climate) misinformation. This analysis focuses on two key facets of user behavior: Tendency \& Persistence and Trend.

\subsection{User Matching}

We used the collected timeline data for the 8,776 users and the classified \politics~and \climate~tweets (described in Section~\ref{sec:data_collection_clasification} and corresponding to Phase 3 in Figure~\ref{fig:data_pipeline}). As the first step, we removed the bots using Botometer~\cite{Yang_Varol_Hui_Menczer_2020}, after which we were left with 7,890 users. We used the classified \politics~and \climate~tweets to segregate the users as \treatment~and \control~users. A user was assigned the \treatment~group if -- \textbf{1)} the user had $\geq 3$ \climate~misinformative tweets and $\geq 0.0012$ as the ratio of \climate~misinformative tweets to the total tweets \textbf{or 2)} the user had $\geq 14$ \politics~misinformative tweets and $\geq 0.0107$ as the ratio of \politics~misinformative tweets to the total tweets. 
In the absence of any past work studying \precovid~misinformation and high variability in the distribution of \politics~and \climate~misinformative tweets, these thresholds were determined by utilizing the median values inherent to the respective variables. 

\begin{figure*}[ht]
    \centering
    \begin{minipage}{0.45\textwidth}
        \centering
        \includegraphics[width=0.6\linewidth]{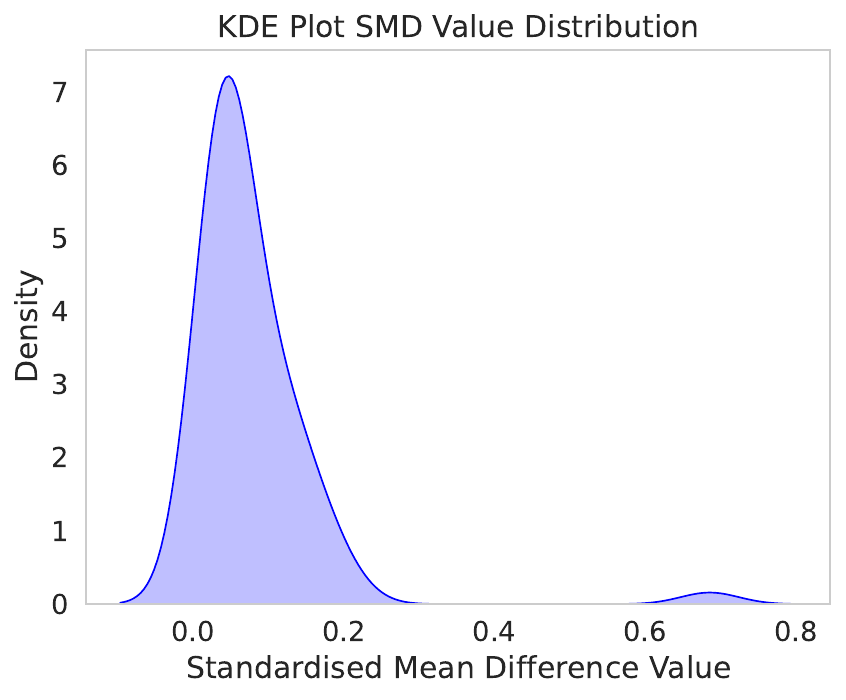}
        \caption{Kernel distribution plot of SMD values of co-variates in matching \control~and \treatment~users.}
        \label{fig:SMD}
    \end{minipage}\hfill
    \begin{minipage}{0.45\textwidth}
        \centering
        \includegraphics[width=0.6\linewidth]{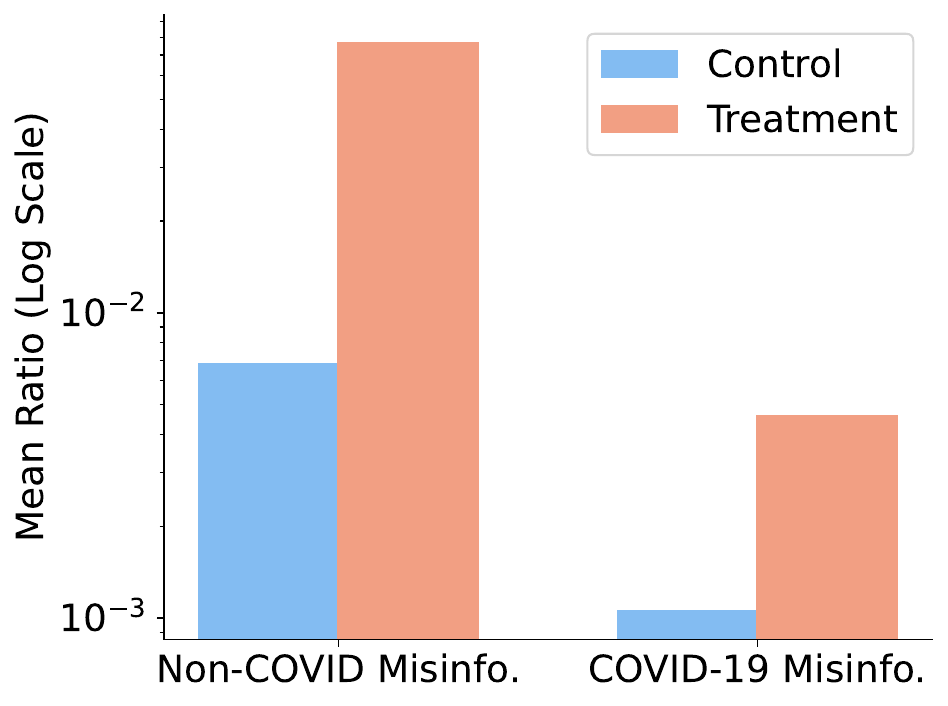}
        \caption{Tendency to spread Non-COVID and COVID-19  misinformation among \control~and \treatment~users 
  during \pericovid~era.}
        \label{fig:rq_2_tendency_bar_plot}
    \end{minipage}
\end{figure*}

Our study follows the framework of causal inference, for which we created the \control~and \treatment~groups. Due to the observational nature of our data, using Randomized Controlled Trials (RCTs) for matching users was impractical and unethical. Hence, we used propensity score matching. We matched the \control~and \treatment~users based on the following co-variates: 1) total tweet count in the \precovid~era, 2) month-wise normalized reply, retweet, like, quote, and tweet count (Jan 2019 - Jan 2020). The selection of these covariates was driven by two primary considerations. Firstly, these covariates have been previously employed in studies for robust propensity score matching~\cite{verma2022examining}. Secondly, the retrospective collection of other covariates, including user follower count, following count, account creation date, average status count, etc., was not feasible utilizing the Twitter API. For matching, we first trained a Logistic Regression classifier predicting the propensity score for each user based on the mentioned co-variates. Next, we paired each \treatment~ user with the most similar \control~ user using the K-nearest neighbor algorithm, keeping a caliper distance $\leq$ 0.2 to ensure a tight matching. This way, we obtained a subset of 5,938 matched users (2,969 users from each set), as shown in Phase 3 of Figure~\ref{fig:data_pipeline}. The Standardized Mean Score (SMD) between the Treatment and Control for each co-variate is less than 0.2 (ref. Figure~\ref{fig:SMD}), indicating tight matching between the users in the two sets.

\subsection{Tendency \& Persistence of Sharing Misinformation} 

First, within RQ2, we sought to understand whether Treatment users are more susceptible to sharing Non-COVID and COVID-19 misinformation during the \pericovid~era. We computed the Non-COVID and COVID-19 misinformation ratio to the total timeline tweets for each user during the \pericovid~time. Figure~\ref{fig:rq_2_tendency_bar_plot} presents the mean ratio of Non-COVID and COVID-19 misinformation shared by the \control~and \treatment~users, respectively. Here, the \treatment~users are $\sim11.22$ (Mann-Whitney U Test;\pval~$<$ 0.005) times more likely to spread the Non-COVID misinformation on average. Similarly, Treatment users are $\sim4.37$ (Mann-Whitney U Test;\pval~$<$ 0.005) times more likely to spread COVID-19 misinformation.

\begin{figure}[h]
    \centering
    \includegraphics[width=0.6\columnwidth]{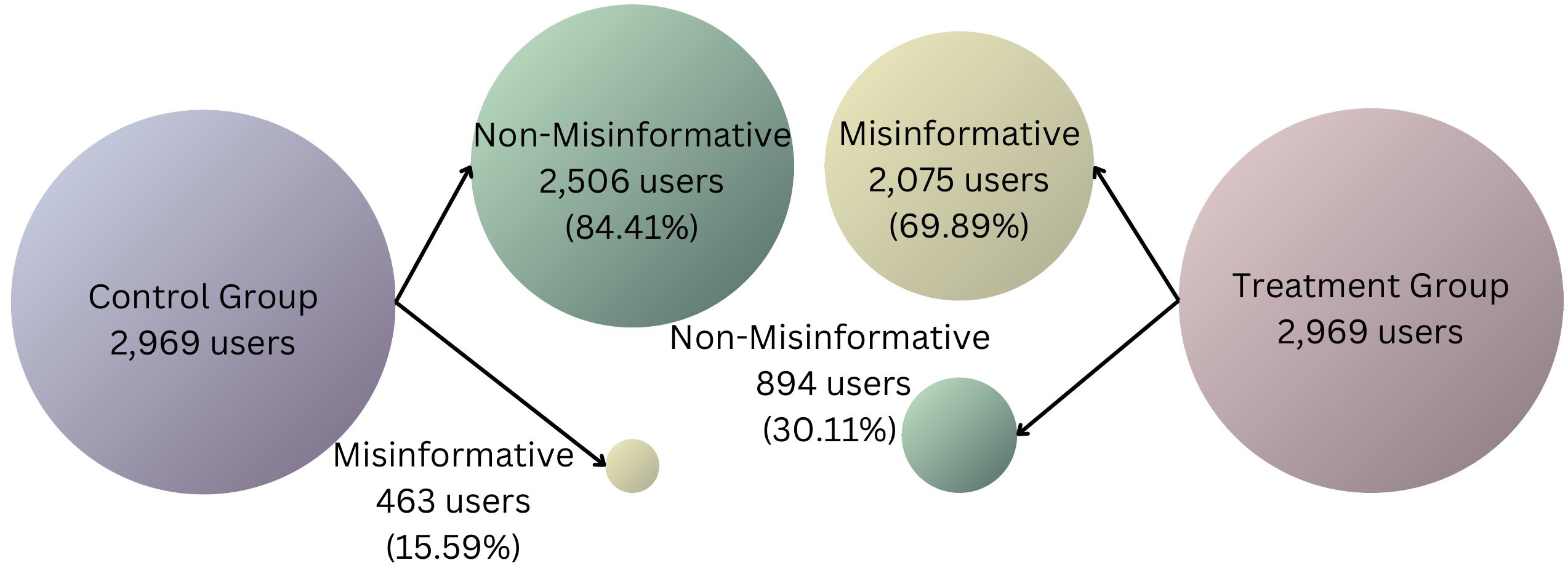}
    \caption{Persistence in behavior to spread COVID-19 misinformation among the \control~and \treatment~users during the \pericovid~era. Bubble sizes are proportional to \#users.} 
    \label{fig:rq_2_persistence_diagram}
\end{figure}

Next, we analyzed the change in the persistence to spread misinformation among the Treatment and Control users before and during the \pericovid~era. Figure~\ref{fig:rq_2_persistence_diagram} presents the distribution of users in the Control and Treatment groups among the \misinfo~and \nonmisinfo~users during the \pericovid~era. As observed, most users preserve their tendency to spread misinformation. 2,075 out of 2,969 ($\sim 70\%$) Treatment users were classified as COVID-19 \misinfo~user. Whereas, 2,506 out of 2,969 ($\sim 84\%$) Control users were classified as COVID-19 \nonmisinfo~user. This results shows that majority of the users preserved their tendency about spreading/not spreading misinformation during the \pericovid~era, even though the domain of misinformation of interest changed from Non-COVID (\climate~and \politics) to COVID-19. Our finding provides a quantitative validation to past works building upon resistance to attitude change theory in the context of online misinformation~\cite{ecker2022psychological}. 

\begin{figure}[h]
    \centering
        \includegraphics[width=0.47\columnwidth]{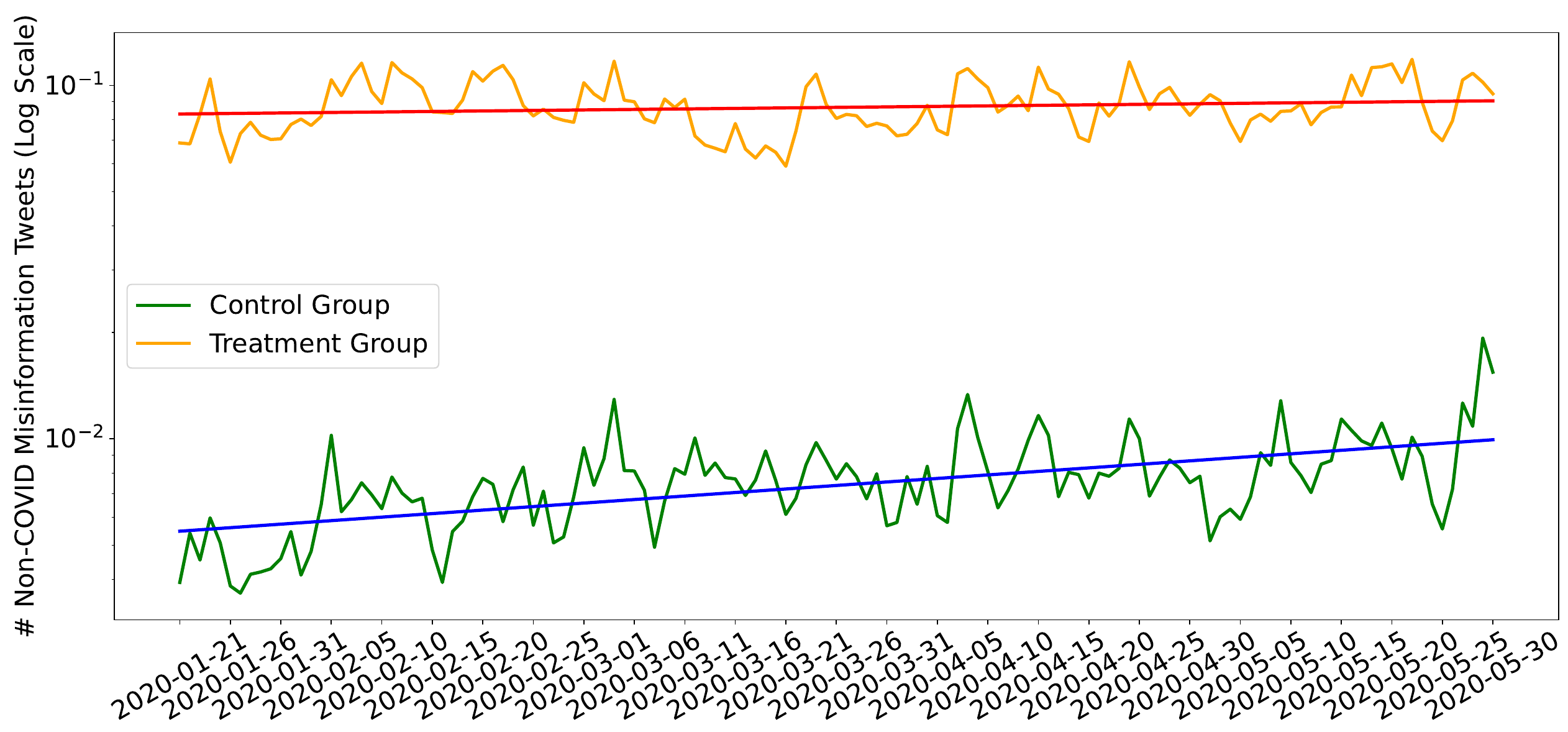}
       \includegraphics[width=0.47\columnwidth]{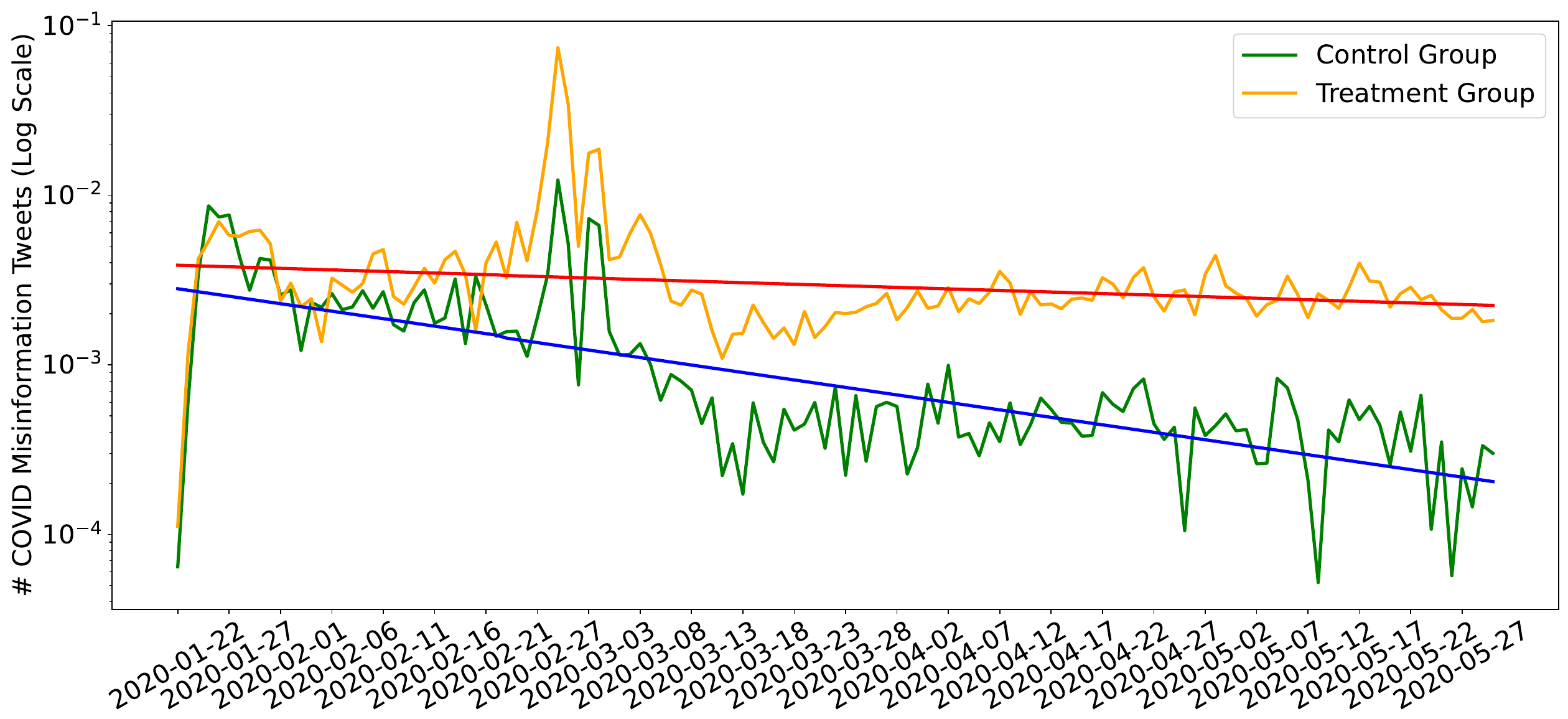}
    \caption{Non COVID (\politics~and \climate) (left) and COVID-19 misinformation (right) sharing trends for \control~and \treatment~group users.}
    \label{fig:rq2_trend_analysis}
    \vspace{-.15in}
\end{figure}

\subsection{Trend of Sharing Misinformation}  


Finally, for RQ2, we aimed to investigate the `spillover' impact of the COVID-19 pandemic on the dissemination of misinformation in tweets related to \climate~and \politics. Through this study, our objective was to gain insights into the broader behavioral shifts of users resulting from a significant event like COVID-19. 
To this end, we conducted a comprehensive analysis of posting trends to analyze the impact of COVID-19 as an event on both the Control and Treatment groups. We calculated the daily proportion of Non-COVID and COVID-19 misinformative tweets among the total tweets posted by users in each group.


Figure~\ref{fig:rq2_trend_analysis} presents the daily activity trends during the \pericovid~period for the \treatment~and \control~groups concerning Non-COVID (left) and COVID-19 misinformation (right). The mean per-day proportion of Non-COVID misinformation for the users in the \treatment~group was $\sim0.088$, whereas the same for the \control~group was $\sim0.008$. Notably, users in the \treatment~group exhibited a significantly higher likelihood of disseminating Non-COVID misinformation, approximately 11 times more than their counterparts (Mann-Whitney U Test; \pval~$<0.001$). Regarding COVID-19 misinformation, the average per-day proportion for the \treatment~group was $\sim0.004$, contrasting with the \control~group's ratio of $\sim0.001$ (Mann-Whitney U Test; \pval~$<0.001$). Overall, we observe an increasing trend in the spread of Non-COVID misinformation among both groups, with the relative increase being higher among the \control~group users. In contrast, we find a general pattern of decline among the COVID-19 misinformative tweets (except for the two weeks in March 2020, when COVID-19 was declared a pandemic). We also observed a steeper decline in the proportion of COVID-19 misinformation tweets among the \control~users than the \treatment~group users.

To elucidate further, we used the spline interpolation method to  analyze the change in the trend of Non-COVID and COVID-19 misinformation ratio for \control~and \treatment~group users. Specifically, we applied linear regression models to each group's data (depicted as linear regression lines in Figure~\ref{fig:rq2_trend_analysis}). For the Non-COVID misinformation, we observed a notable difference in the regression coefficient between the \treatment~and \control~group regression models, with the \control~group's model having $\sim6.93$ times higher coefficient. In contrast, the Treatment group's regression model intercept was $\sim6.1$ times higher than that of the Control group. In the context of COVID-19 misinformation, a more expected trend emerged, with both groups displaying regression coefficients below zero. We observed that the rate of decline in COVID-19 misinformation among \control~group users was nearly 4.8 times more rapid than that among \treatment~group users. 

The findings of this analysis show that \treatment~users showed significantly higher chances of sharing both COVID-19 and Non-COVID misinformation than their counterparts. Moreover, our analysis shed light on the ``spillover effect'' of the pandemic on cross-domain misinformation, where users from both groups showed an increasing inclination towards sharing Non-COVID misinformation with time. In contrast, we observed a decline in the dissemination of COVID-19 misinformation over time -- a trend consistent with previous research findings~\cite{shahi2021exploratory}. Additionally, our analysis aligns with prior studies that have identified a surge in political misinformation during the COVID-19 pandemic~\cite{jungkunz2021political}. Finally, the onset of the COVID-19 pandemic caused a notable shift in collective attention, diverting focus from prevalent domains such as climate, public events, and electoral politics to the pervasive topic of COVID-19~\cite{he2021identifying, leng2021misinformation}. This shift in attention may have contributed to an escalation in the dissemination of non-COVID-19-related misinformation.

%% file: sections/rq3_liwc_analysis.tex
Finally, we focus on comprehending the influence of COVID-19 as an event on both \treatment~and \control~users' psychological behavior. 
Specifically, we analyze the tweets discussing content unrelated to COVID-19 for users belonging to both the \treatment~and \control~groups during the \pericovid~as compared to \precovid~era. For this task, we used Linguistic Inquiry and Word Count (LIWC)~\cite{pennebaker2001linguistic} to uncover the linguistic characteristics. LIWC is an established psycholinguistic lexicon that has been shown to effectively explain the psychological behavior of users using text data~\cite{10.1145/3274351, saha2020causal}. Employing a Difference-in-Difference (DID) analysis approach~\cite{angrist2009mostly}, we analyze the relative psychological effect of COVID-19 on \control~and \treatment~group users.

\subsection{Approach}
We used tweets from \control~and \treatment~groups encompassing topics unrelated to COVID-19 to examine the changes induced by the pandemic on the psychological behavior of users in both groups pertaining to topics unrelated to the pandemic. The timeline tweets for each group were categorized into \precovid~and \pericovid~segments, similar to the methodology employed in addressing \textbf{RQ2}. For methodological consistency, we applied a series of text pre-processing steps, including lowercasing, lemmatization, and removal of punctuation and commonly occurring text such as `RT.' For each temporal segment, we computed the LIWC scores across all categories for each pre-processed tweet, followed by user-level averaging. 
Then, we normalized the per-user LIWC score across each of the \control~and \treatment~groups. 
We computed the DID values through subtracting the change in score for each category among the \treatment~group (\pericovid~- \precovid) from \control~group (\pericovid~- \precovid). 

\begin{table}[ht]
    \small 
    \centering
    \begin{tabular}{|l|p{0.14\columnwidth}|p{0.14\columnwidth}|p{0.12\columnwidth}|p{0.15\columnwidth}|}
    \hline
    \textbf{Category} & \textbf{Tr Mean} 
    & \textbf{Ct Mean} 
    & \textbf{DID} & $p$ \textbf{value}\\
    \hline
    \rowcolor{blue!10} verb & 1.582 & -3.424 & 5.006 & $<10^{-15}$ \\
    \rowcolor{blue!10} social & 4.900 & 0.022 & 4.878 & $<10^{-15}$ \\
    \rowcolor{blue!10} conj & 1.044 & -0.847 & 1.891 & $<10^{-15}$ \\
    \rowcolor{blue!10} informal & 1.831 & 0.007 & 1.824 & $<10^{-15}$ \\
    \rowcolor{blue!10} adverb & 0.570 & -1.043 & 1.612 & $<10^{-15}$ \\
    \rowcolor{blue!10} focuspresent & 1.029 & -0.332 & 1.361 & $<10^{-15}$ \\
    \rowcolor{blue!10} anger & -0.201 & -1.344 & 1.143  & $<10^{-15}$ \\
    \rowcolor{blue!10} posemo & 0.817 & -0.190 & 1.008 & $<10^{-15}$ \\
    \rowcolor{blue!10} anx & 0.682 & -0.301 & 0.983 & $<10^{-15}$ \\
    \rowcolor{blue!10} focusfuture & 0.121 & -0.551 & 0.672 & $<10^{-15}$ \\
    \hline
    \hline
    \rowcolor{red!10} netspeak & -5.981 & 0.008 & -5.989 & $<10^{-15}$ \\
    \rowcolor{red!10} function & 0.340 & 4.819 & -4.479 & $<10^{-15}$ \\
    \rowcolor{red!10} nonflu & -1.907 & -0.063 & -1.844 & $<10^{-15}$ \\
    \rowcolor{red!10} relativ & -1.186 & 0.539 & -1.726 & $<10^{-15}$ \\
    \rowcolor{red!10} cogproc & -0.606 & 0.962 & -1.568 & $<10^{-15}$ \\
    \rowcolor{red!10} ppron & -1.489 & -0.120 & -1.369  & $<10^{-15}$ \\
    \rowcolor{red!10} we & -1.155 & 0.013 & -1.168 & $<10^{-15}$ \\
    \rowcolor{red!10} article & -0.053 & 0.630 & -0.683 & $<10^{-15}$ \\
    \rowcolor{red!10} assent & -0.631 & 0.014 & -0.644  & $<10^{-15}$ \\
    \rowcolor{red!10} affect & -0.594 & 0.011 & -0.605 & $<10^{-15}$ \\
    \hline
    \end{tabular}
    \caption{Top 10 positive DID and top 10 negative DID across LIWC categories. Tr is \treatment~ and Ct is \control~ group.}
    \label{tab:did_analysis_2}
    \vspace{-1.5em}
\end{table}

\subsection{Psycholinguistic Differences}

Broadly, we found that 68 (out of 73) LIWC categories showed distinctive linguistic behavior among both groups, based on Holm-Bonferroni correction.
Table~\ref{tab:did_analysis_2} presents the top LIWC score changes for both groups across the \textit{Pre} and \pericovid~era, along with the DID values. 

More elaborately, \treatment~users increased their use of verbs (`verb' category) from \precovid~to \pericovid~(Tr mean change: 1.582 (2.091-0.509)), whereas the usage among the \control~users decreased (Ct mean change: -3.424 (1.481-4.905)). This suggests that the \treatment~users aimed for a more dynamic and attention-grabbing communication style during the \pericovid~era, potentially to evoke emotional responses through their content. This aligns with research on persuasive communication, suggesting that strategic action-oriented language enhances message persuasiveness \cite{kurvers2021strategic}. Similarly, \treatment~users increased the usage of positive emotion in the text (`posemo' category) (Tr mean change: 0.817 (1.431-0.613)). In contrast, the usage decreased among the \control~users (Ct mean change: -0.190 (0.483-0.673)). While the negative impact of the pandemic can explain the decrease in positive emotions among the \control~users, past research linking emotions and deception has shown that usage of positive emotion language increases gullibility and decreases the ability of users to detect deception, ultimately influencing their behavior~\cite{east2007mood}.

Contrary to the above two categories, we observed a large increase in the use of informal language in the \treatment~group, while the \control~users kept their behavior similar during the \pericovid~era as compared to the \precovid~times (\treatment~Mean: 1.831, \control~Mean: 0.007). Research studying misinformation during COVID-19 has shown higher usage of informal language among users spreading misinformation~\cite{memon2020characterizing}. Supporting the involvement in spreading misinformation, the \treatment~group users also exhibited a higher decrease in personal pronoun usage (`ppron' category) as compared to \control~group (Tr mean change: -1.489 (0.459-1.948), Ct mean change: -0.120 (0.868-0.988)). \treatment~users opted for a less self-centric approach, perhaps to appear as objective sources of information. 
It is known that self-distancing can be a strategy adopted to minimize personal involvement in deceptive messages \cite{lying2003Newman}. 

The \treatment~group also exhibited a notable shift in their language usage towards a heightened focus on the present moment, as evident in their increased usage of the `focuspresent' category (Tr mean change: 1.029 (1.399-0.370) while the \control~group displayed a contrasting trend (Ct mean change: -0.332 (1.035-1.367)). On manual validation, we observed that the \treatment~group users were more inclined to react and comment on current events in an attempt to create urgency or higher emotional engagement. While the usage of `function' words among the \treatment~group users remained similar(change in \treatment~Mean: 0.340 (1.003-0.663)), the usage significantly increased among the \control~group users (change in \control~Mean: 4.819 (5.007-0.189)). The \control~group's significantly higher usage of functional words in the \pericovid~era in comparison to \precovid~times indicates a deliberate attempt toward a clear, structured, and informative communication, possibly prioritizing comprehensive content delivery. Finally, we observed lower usage of `assent' category words among the \treatment~group users, whereas the usage among the \control~group users slightly increased during the \pericovid~era in comparison to the \precovid~time period. Decreased usage shows that the \treatment~users became less agreeable to opinions that did not align with their ideologies during the pandemic. Moreover, adopting a contrarian communication style has been shown to increase engagement~\cite{kurvers2021strategic}. The observed lack of significant change within the `assent' category usage from the \control~group users indicates a group seemingly uninterested in grabbing attention and focusing on conveying facts.

Through this analysis, we showed that COVID-19 had a significant impact on the psychological behavior of both \control~and \treatment~group users. We observed a `spillover' effect of the pandemic over how users discussed about topics unrelated to COVID-19. 
Our findings highlight the deliberate effort by the \treatment~group users to indulge in spreading Non-COVID misinformation. 
In contrast, \control~group preferred a more responsible and controlled approach while discussing topics apart from COVID-19.

%% file: sections/discussion.tex

\subsection{Implications for Tackling Online Misinformation}

\subsubsection{An Ecological Approach to Studying and Countering Online Misinformation}

In contrast to the majority of the existing work that focuses on addressing the misinformation pertaining to a specific domain~\cite{ruchansky2017csi}, we presented a more holistic approach spanning an array of three domains. As observed in our study, while the tendency for propagating Non-COVID-19-related misinformation exhibited an upward trajectory during the pandemic, the tendency for disseminating COVID-19 misinformation demonstrated a declining pattern, regardless of people's historical misinformation spreading behaviors.
These results give credence to previous works that found increased levels of political polarization and misinformation during COVID-19~\cite{jungkunz2021political}. Moreover, our findings align with previous research studying collective attention shifts within the public sphere during major events, exemplified by heightened discussions on relevant entities related to the focal event of interest~\cite{lin2014rising, wakamiya2015portraying}. The COVID-19 pandemic induced a significant shift in collective attention, transitioning from prevalent topics such as \climate,~public events, and electoral politics to COVID-19~\cite{he2021identifying, leng2021misinformation}. This may have led to information overload, and, per media dependency theory, promoted higher sharing of unverified Non-COVID misinformation among the Control and Treatment users~\cite{laato2020drives, ball1976dependency}.

Additionally, previous studies have also shown a decline in COVID-19 misinformation with time due to reduced ambiguity and counter-misinformation efforts~\cite{evanega2020coronavirus,shahi2021exploratory} -- a finding also corroborated by our study, but richly situated in the backdrop of misinformation sharing behaviors on other topics~\cite{bessi2015trend}. Hence, the distinctive patterns observed across the three domains in our study underscore the significant variation in the trend of misinformation spread, echoing past research emphasizing domain-specific nuances~\cite{bessi2015trend}. In light of these intricate dynamics, our study highlights the necessity of adopting an ecological approach to address cross-domain misinformation. Such an approach, cognizant of the multifaceted media landscape, ensures that countering strategies transcend specific information types, allowing the newer strategies to navigate the complexities of each domain and issue within the broader context of others.

\subsubsection{Anchoring on the Humans Behind Misinformation}

As noted earlier, a substantial portion of online misinformation tends to stem from a small group of users, making user-based intervention techniques pragmatic and essential~\cite{ccdhreport, qian-etal-2018-leveraging}. Hence, to effectively design these interventions, understanding the psychology behind \textit{why} individuals share misinformation is crucial~\cite{ecker2022psychological} -- a key motivation of our work. After all, in 2016, the Oxford Dictionary chose ``post-truth'' as the word of the year\footnote{https://languages.oup.com/word-of-the-year/2016/}, defined as pertaining to situations where appeals to emotion and personal belief have a greater impact on shaping public opinion than objective facts.

Indeed, in our study, we observe that 
distinct sets of external stimuli, for instance, the devastating COVID-19 pandemic, exerted varying effects on individuals, contingent upon their inclination to disseminate misinformation in the past. These differences affected users' posting patterns and their psychological behavior, ultimately shaping the propagation dynamics of misinformation. These findings about the correlative relationship between historical inclination towards sharing misinformation and current behavior confirm observations in prior research that have shown that inherent bias and strong conspiratorial tendencies motivate individuals to spread misinformation~\cite{hornsey2017attitude}.

In a related context, it has been observed that social media platforms play a significant role in both fostering the establishment and fortification of false beliefs within the user community~\cite{jamieson2008echo,del2016spreading}. Such users have also been documented to exhibit a proclivity for transitioning between diverse topics. Aligning with the outcomes of our study, past research indicates that the belief in false information on a particular subject positively influences the prevalence of misinformation on other topics~\cite{bessi2015trend}. Therefore, equipped with an understanding of the psychological drivers and behaviors of people with a propensity to share misinformation, it may be possible to design novel human-centered intervention strategies. 


\subsection{Practical Strategies and Solutions}


\subsubsection{Agile strategies}

Based on our findings, we propose that continuous and adaptive fact-checking mechanisms and collaborations between tech companies, researchers, and news organizations should be adopted to identify and combat misinformation effectively. An ecological strategy that considers misinformation as a multifaceted problem spanning topics, issues, and events can allow for the use of fact-checking and correction in some contexts, algorithmic promotion of trusted sources in some, media literacy education in some others, while considering enforcing legal frameworks to hold individuals or organizations accountable elsewhere.

It is important that our findings inspire approaches to promote media literacy and critical thinking, alongside inoculation efforts. Sometimes described as an ``arms race''\footnote{https://www.pewresearch.org/internet/2017/10/19/the-future-of-truth-and-misinformation-online/}, we note that as the counter-misinformation strategies become more sophisticated, those spreading misinformation may also adapt and find new ways to evade detection or counter the interventional efforts. Thus our findings should be interpreted and used in a way that adopts a collaborative approach to countering online misinformation, rather than villifying or victimizing certain individuals.

\subsubsection{Strategic use of reactive and preemptive interventions} 

Studies have demonstrated the efficacy of both preemptive and reactive strategies in the mitigation of online misinformation, depending upon the specific contextual circumstances in which they are employed~\cite{vraga2020testing}. As observed in our study, \misinfo~users started disseminating misinformation related to COVID-19 in the early stages of the pandemic, and they sustained their behavior over an extended period. Hence, reactive strategies  focused on algorithmically boosting information from credible sources pertaining to specific content (rather than generalized topics) during the early phases of significant events can help~\cite{ecker2022psychological}. Further, the intricate association between users’ historical misinformative behaviors online and their present actions on social media highlights the importance of using user-centric preemptive strategies. Most effective preemptive strategies have been derived from the inoculation theory~\cite{compton2021inoculation}. Strategies that focus on first warning users about misinformative content and then identifying the techniques used for propagating misinformation have been shown to be effective in equipping users against a wide array of topics in the future.

Our study is centered on the examination of user behavior, with a specific focus on investigating the relationship between users' historical tendencies in disseminating misinformation and their present behavior in this regard. Consequently, we recognize that there exists a conceivable risk of misusing the insights from our work if users on social media platforms are subject to biased tracking or inoculation-based interventions solely based on their historical behaviors, particularly regarding the anticipation of future misinformation sharing. The act of deplatforming users predicated on such anticipations has the potential to limit users' freedom of expression~\cite{lewandowsky2020technology}. Moreover, preemptively moderating users assuming they might share misinformation in the future may have chilling effects on speech, negatively impacting the online ``public sphere'' that may social media platforms aspire to be. Preemptive user moderation can also disempower individuals from developing critical thinking skills and media literacy~\cite{nathanson2002unintended}.

\subsubsection{Empathetic solutions.} Our findings encourage us to adopt a more compassionate and constructive approach to addressing the issue. Rather than vilifying individuals, we can focus on empowering them with the skills and knowledge needed to discern reliable sources and critically evaluate information. This approach could not only reduce the stigma associated with sharing misinformation but also promote a more inclusive and empathetic digital environment. Furthermore, those with a history of susceptibility to misinformation could be discouraged from engaging as novel topics or issues emerge. Instead they could be algorithmically nudged to pursue accurate information rather than seeking evidence that aligns with their pre-existing biases~\cite{scheufele2019science}.

\subsubsection{Cautious implementation of measures.}

We note that we cannot ignore potential counterproductive reactions resulting from following some of the practical strategies and solutions we propose in this paper. Some research and media advocates suggest that directly confronting people with information that contradicts their beliefs can lead to a backfire effect where they become more entrenched in their views~\cite{swire2020they}. Such corrections can often be perceived as an assault on one's identity, triggering a cascade of evaluations and emotional responses that impede the process of information revision~\cite{hornsey2017attitude}. Hence, strategies promoting corrections while providing identity affirmations to the recipient could prevent this backfire effect~\cite{paynter2019evaluation}.

A recurrent challenge confronting social media platforms involves the nuanced task of defining misinformation~\cite{bednar2008bias}. Often, the definitions are subjective introducing a potential risk of bias and ultimately resulting in selective enforcement of bad actors. Expanding upon this concern, there is a risk that inoculation efforts based on users' historical misinformation-sharing behaviors, as investigated in this paper, could be misused selectively to target certain viewpoints or political ideologies, leading to accusations of bias and censorship. Hence, platforms need to proceed with caution while implementing newer measures to address the issue related to online misinformation. Lastly, it is crucial to note that intervention strategies could yield counterproductive reactions if they lack transparency and fairness. The effectiveness of these measures hinges on ensuring clear communication and fairness in their implementation~\cite{vaccaro2020end}. Moreover, previous research has demonstrated that the efficacy of corrective measures is enhanced when detailed refutations are provided, as opposed to simple retractions that lack explanatory information on why the misinformation is inaccurate~\cite{swire2017role}. 
This underscores the imperative for platforms to develop intervention algorithms that prioritize transparency and fairness, mitigating the risk of unintended biases and ensuring equitable corrective measures.



\section{Conclusion, Limitations, and Future Work}
\label{sec:conclusion_limitation}

Our society today desperately needs resilient solutions to online misinformation that targets its root causes, rather than piecemeal strategies that shift attention to detecting misinformation on whatever might be the biggest threat to our information ecosystems at the moment. This will not be possible without a nuanced and intricate understanding of the humans who share such content as well as the temporal dynamics that interweave misinformation sharing across diverse topics and issues. Specifically, our study highlights distinct posting patterns and psychological drivers contrasting users who share misinformation and those who do not. Additionally, we observed an intricate link between users' past misinformation-sharing behavior and their present actions. Through this work, we contribute to expanding our knowledge of potential solutions.

Although novel, our study has a few limitations. Firstly, despite our rigorous efforts to minimize sampling bias in our data collection methodology, it is crucial to acknowledge that the analysis is based on a sample of the data accessible on the platform, totaling over 61 million tweets. Additionally, the causal framework employed in our investigation was implemented within the constraints of our study, wherein data pertaining to a subset of 5,938 users was analyzed. This limitation arose from the restrictions imposed by the Twitter API, which imposes a rate limit on the retrieval of historical tweets, allowing access to only 10 million such tweets per month. Furthermore, our data collection efforts were hindered by the cessation of free academic access to the Twitter API, rendering it unfeasible for us to gather timeline data for a larger cohort of users. We also recognize the inherent limitations of the quasi-experimental study design implemented in this work, as it does not establish definitive causality. However, our study design ensures more robustness than a correlational analysis by effectively minimizing the influence of confounding variables. Hence, the insights derived from this study can motivate future computational studies focused on analyzing human behavior in the context of spreading misinformation. Moreover, these findings can contribute to a deeper understanding of misinformation propagation in cross-domain settings, thereby inspiring the development of more comprehensive strategies for countering misinformation.

Next, the scope of our study was confined to the examination of topics related to \politics,~and~\climate, since comprehensive, high-quality publicly accessible datasets pertinent to other domains were unavailable. Given the observational nature of our research, we refrained from starting with the data collection and annotation process encompassing diverse topics, such as celebrity gossip and general knowledge facts. That said, an intriguing prospect for future research resides in the longitudinal analysis of cross-domain misinformation dynamics, particularly within perennial topics like \politics and \climate, juxtaposed with transient topics such as controversies involving public figures and epidemic occurrences.

Finally, it is important to recognize the role played by platform algorithms and social influence in shaping user behavior~\cite{posetti2018news,kozyreva2020citizens}. However, this aspect was beyond the scope of the present study and represents a potential avenue for future research exploration.